\documentclass[runningheads,a4paper]{llncs}

\usepackage{times}
\usepackage{amssymb}
\usepackage{graphicx}
\usepackage{url}

\usepackage{amssymb}
\usepackage{multicol}
\usepackage[table,xcdraw]{xcolor}
\usepackage{pifont}
\usepackage[hidelinks]{hyperref}
\usepackage{caption}      
\usepackage{capt-of}      

\newcommand{\cmark}{\ding{51}}%
\newcommand{\xmark}{\ding{55}}%
\def\weblink#1{\href{#1}{#1}}

\newcommand{\colorize}[3]{\cellcolor[HTML]{#1}{\color[HTML]{#2}#3}}

\newcommand{\keywords}[1]{\par\addvspace\baselineskip
\noindent\keywordname\enspace\ignorespaces#1}

\urldef{\mailsa}\path|xzelina2@fi.muni.cz|

\begin{document}

\title{Computing patient similarity based on unstructured clinical notes}

\titlerunning{Patient similarity based on unstructured clinical notes}

\author{Petr Zelina\inst{1} \and Marko Řeháček\inst{1} \and Jana Halámková\inst{2,3} \and Lucia Bohovicová\inst{2} \and\newline Martin Rusinko\inst{2,3} \and Vít Nováček\inst{1,4}}


\authorrunning{Petr Zelina et al.}

\institute{$^1$Faculty of Informatics, Masaryk University \\
\inst{2}Department of Comprehensive Cancer Care, Masaryk Memorial Cancer Institute \\
\inst{3}Faculty of Medicine, Masaryk University \\
\inst{4}Bioinformatics research group, Masaryk Memorial Cancer Institute \\
Corresponding author: \href{mailto:xzelina2@fi.muni.cz}{xzelina2@fi.muni.cz}}

\index{Zelina, Petr}
\index{Řeháček, Marko}
\index{Halámková, Jana}
\index{Bohovicová, Lucia}
\index{Rusinko, Martin}
\index{Nováček, Vít}

\toctitle{} \tocauthor{}

\maketitle

%
%
%
%
\begin{abstract}
Clinical notes hold rich yet unstructured details about diagnoses, treatments, and outcomes that are vital to precision medicine but hard to exploit at scale. We introduce a method that represents each patient as a matrix built from aggregated embeddings of all their notes, enabling robust patient similarity computation based on their latent low-rank representations. 
Using clinical notes of 4,267 Czech breast-cancer patients and expert similarity labels from Masaryk Memorial Cancer Institute, we evaluate several matrix-based similarity measures and analyze their strengths and limitations across different similarity facets, such as clinical history, treatment, and adverse events. The results demonstrate the usefulness of the presented method for downstream tasks, such as personalized therapy recommendations or toxicity warnings.
\vspace{-0.8\baselineskip}
\end{abstract}


%
%
%

\section*{Introduction}
\vspace{-0.7\baselineskip}
Electronic health records (EHRs) are digital artifacts with which healthcare pro\-vi\-ders store, integrate, process and access all information associated with a patient's journey through the medical system.
Over the last couple of decades, EHRs have been utilized  to gain insights from retrospective data and, more recently, to facilitate clinical decision support~\cite{rajpurkar2022ai}. Examples of such approaches include assessing stroke severity~\cite{kogan2020assessing} or predicting lung cancer relapse~\cite{janik2023machine}. One of the common challenges reported by virtually every work in this domain is the scarcity of structured EHR data. This is most often mitigated by extracting relevant facts from the unstructured content of the records (e.g., plain text in clinical notes, which is the modality we focus on in this study).

While modern clinical text mining tools may allow for reliable named entity and relation extraction, this is often limited to very specific contexts or entity/relation ty\-pes~\cite{percha2021modern}. Therefore, a truly comprehensive and accurate extraction of structured information from the plain text parts of EHRs is still largely an open problem. In this study, we explore a complementary angle of processing unstructured patient data in EHRs: representing the patients \textit{en bloc} using the textual parts of their records, and computing patient similarities from the representations. The similarities can then support personalized medicine via various downstream applications, such as survival or response prediction based on the clinical history of similar patients, and corresponding personalized treatment recommendations or potential treatment toxicity alerts.

Regarding the representation of the patients based on the textual elements of their EHRs, we have investigated different embedding techniques -- from the classics like \textit{Latent semantic analysis}~\cite{lsa} to more modern ones like \textit{Doc2Vec}~\cite{doc2vec} and \textit{transformer-based embeddings}~\cite{sentencebert}. The vector representations of individual records are stacked into matrices that fully represent the corresponding patients, and thus can be used for computing patient similarities through matrix similarity scores. We explore several applicable scoring methods in combination with different embedding techniques and analyze the influence of various hyperparameters on their performance. Last but not least, we evaluate the resulting patient similarity measures in a validation study involving three clinical experts.

The main contributions of the presented work are: a method for computing patient similarities based on their clinical notes automatically segmented by a technique introduced in~\cite{zelina2022unsupervised}; a modular experimental pipeline for combining various state-of-the-art text embedding and matrix similarity techniques within the proposed method; a methodology for validating the computed patient similarities across different clinical contexts (e.g., patient medical history, diagnosis or treatment type); a thorough discussion of the results achieved and lessons learned. The source code for the experimental pipeline, including the validation, is available on GitHub\footnote{
\weblink{https://github.com/ZepZep/patient-similarity}
}.

\begin{figure}[h]%
\centering
{\includegraphics[width=0.60\linewidth]{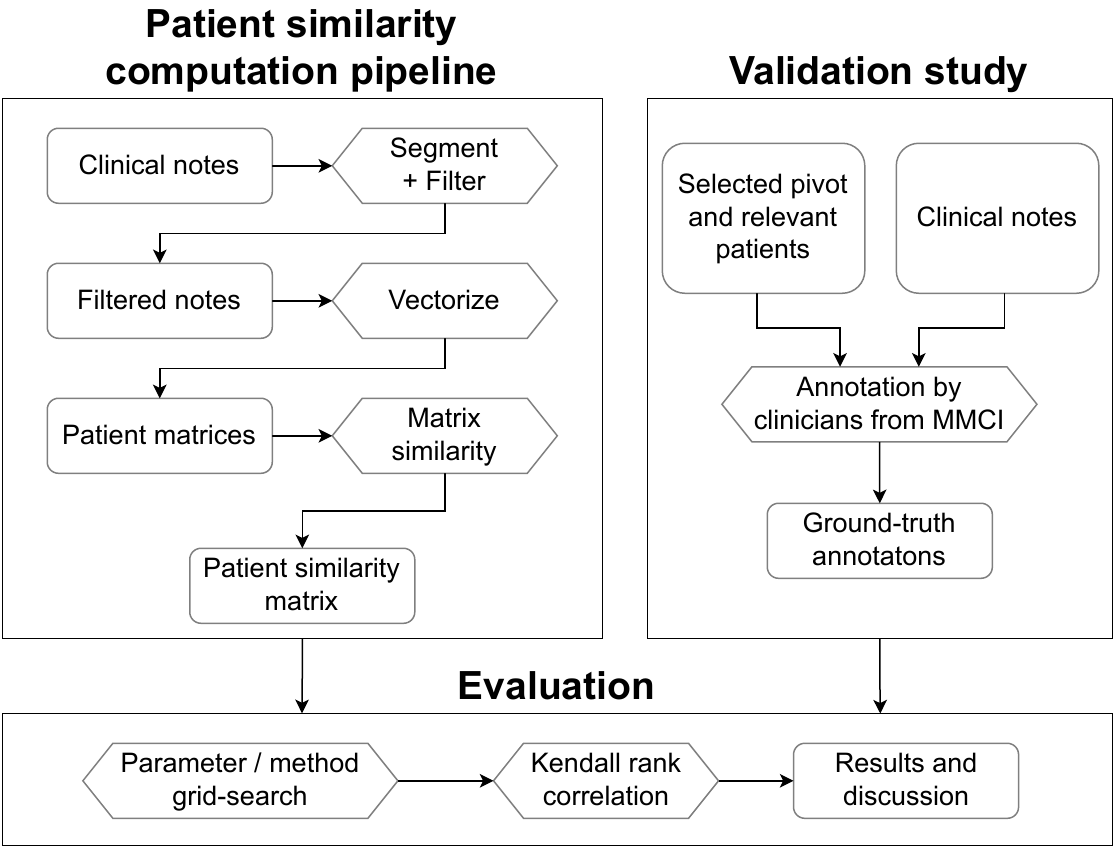}}
\caption{Overview of our methodology. Adapted from~\cite{Zelina2023thesis}.}\label{fig:overview}
\end{figure}

Figure \ref{fig:overview} shows the overall experimental setup for computing patient similarity. In the consequent sections, we describe the experiments in detail and present the validation study design. Then we discuss the results of the experiments, and the impact of various design decisions and hyperparameter values on the performance of each investigated algorithm.

Full details on the implementation of the experiments are available in the thesis~\cite{Zelina2023thesis} on which this substantially extended study is based (please note that the results presented here supersede the ones in the thesis due to an updated experimental pipeline with some major fixes).

\section{Similarity computation pipeline}

In all variants of the experimental settings, the framework studied in this review calculates \textit{absolute} patient similarity -- how similar two patients are in their background (things that do not change much over time, like social history or allergies) and in the current state of their treatment.

\subsubsection{Pipeline overview}
The top-left part of Figure \ref{fig:overview} shows the three steps for calculating patient similarity. The system starts with unstructured clinical notes and outputs a similarity score for each pair of patients.

\begin{itemize}

\item We start with a sequence of \textbf{Clinical notes} for each patient.

\item The first step is \textbf{Segmentation} and \textbf{Note filtering}. Patient similarity can have many different aspects, like \textit{family history}, \textit{treatment} or \textit{comorbidities}. This step splits each clinical note into parts and leaves only the relevant parts for a given similarity category.

\item After filtering, we have a sequence of \textbf{Filtered notes} for each patient.

\item The next step is to \textbf{Vectorize} each filtered note. We investigate 3 different vectorization techniques (\texttt{vmethod}): \textit{LSA} (\texttt{Vlsa},~\cite{lsa}), \textit{Doc2Vec} (\texttt{Vd2v},~\cite{doc2vec}) and \textit{transformer-based embedding} (\texttt{Vrbc},~\cite{sentencebert}). For each of these techniques, one can specify the dimension of the output vectors (\texttt{dim}).

\item The result of the vectorization process is a \textbf{Patient matrix} with dimension \texttt{num-} \texttt{ber\_of\_notes} $\times$ \texttt{dim} for each patient.

\item The final step is to calculate \textbf{Matrix similarity} between each pair of patients (i.e. patient matrices). We experiment with 3 different methods (\texttt{mmethod}): \textit{RV Coefficient} (\texttt{Rrv2},~\cite{rv2}), \textit{MaxMax similarity} (\texttt{Rmms}) and \textit{Edit distance similarity} (\texttt{Reds}).

\item The final result is a single \textbf{Patient similarity matrix} of dimension \texttt{number\_of\_} \texttt{\_patients} $\times$ \texttt{number\_of\_patients}. 

\end{itemize}

\subsubsection{Clinical note segmentation and filtering}
\label{chap:filtering}

The concept of general patient similarity is not well defined, because two patients may be similar, for example, with regards to their \textit{social history} but completely different in terms of their \textit{treatment} or current \textit{comorbidities}. For this reason, we split the similarity into 10 categories based on our consultations with three clinical oncology experts. (see section \ref{chap:val-categories})

The segmentation is based on~\cite{zelina2022unsupervised}. First, we split notes into paragraphs. As 60\% of the segments include titles like \texttt{Medication:} or \texttt{Summary:}, we train a BERT-based title-predictor to label the rest. Finally we create a vector space of the segment titles to measure title similarity.


We devised a semi-automatic approach to determine which segment types (titles) are relevant to which patient similarity category. We select prototype titles for each similarity category and use the latent space to choose similar segment types that are also relevant to the similarity category. 
For example, when selecting relevant segments for the \textit{Medication} similarity category, we chose the \textit{Medication: [medikace:]}  title and used the latent space to discover, for instance, that \textit{Drugs: [léky:]} and \textit{M: [m:]} (abbreviation of medication) segment types contain similar information and thus are relevant as well. One segment type can be relevant to zero, one or many similarity categories.

With these methods and segment type relevancy bitmaps, we can split each clinical note into segments and leave only the relevant segments for a given similarity category. This ensures that the final similarity score better reflects the aspects of the given similarity category.

\begin{figure*}[t]
\centering
\includegraphics[width=0.90\linewidth,]{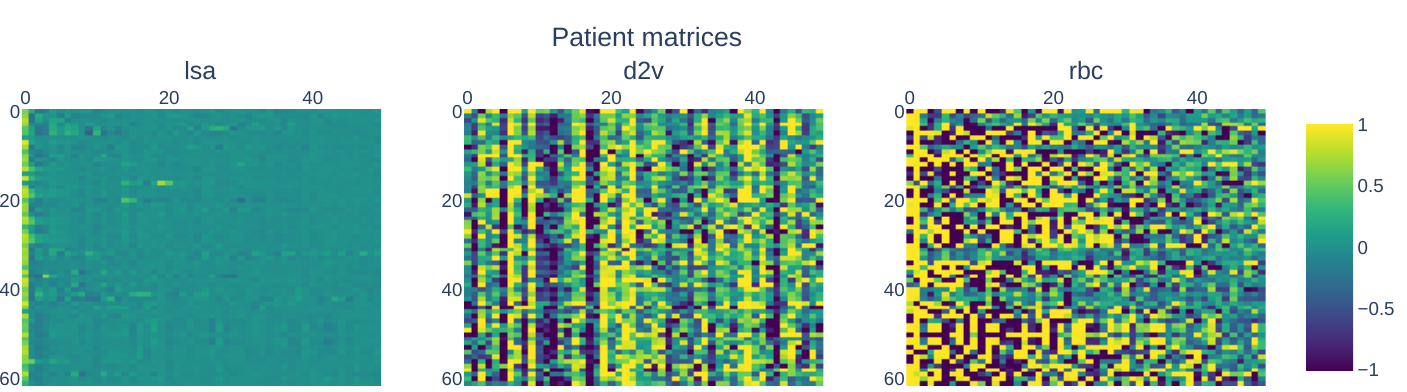}
\caption[Patient matrices]{Example of patient matrices of a patient with 62 clinical notes (rows). The embedding dimension is 50 (columns).~\cite{Zelina2023thesis}}
\label{fig:pacmat}
\end{figure*}

\subsubsection{Clinical note vectorization}\label{chap:note-vectorization}
After creating similarity-category-specific notes in the first step, the second step transforms the notes into vectors. We explore three methods.
\begin{itemize}
    \item Latent semantic analysis (\texttt{Vlsa})~\cite{lsa}.
    \item Doc2Vec embedding (\texttt{Vd2v})~\cite{doc2vec} with the PV-DM algorithm.
    \item Transformer [CLS] embedding (\texttt{Vrbc})~\cite{sentencebert}. We use the RobeCzech model finetuned on the segment classification task with SVD compression.
\end{itemize}

Given a patient $P$ with $n$ clinical notes and a vectorization technique with output dimension $d$, we can represent this patient with a matrix $M_P \in \mathbb{R}^{n \times d}$ by vertically stacking individual embeddings of their notes. The $k$-th row of this matrix represents the $(k+1)$-st patient note in chronological order. Figure \ref{fig:pacmat} shows examples of patient matrices with different vectorization methods.

\subsubsection{Patient matrix similarity}
The final step is to reduce each pair of patient matrices with possibly different numbers of records (rows) into a single similarity. We review three methods for calculating this similarity score.

\begin{figure*}[]%
\centering
{\includegraphics[width=\textwidth]{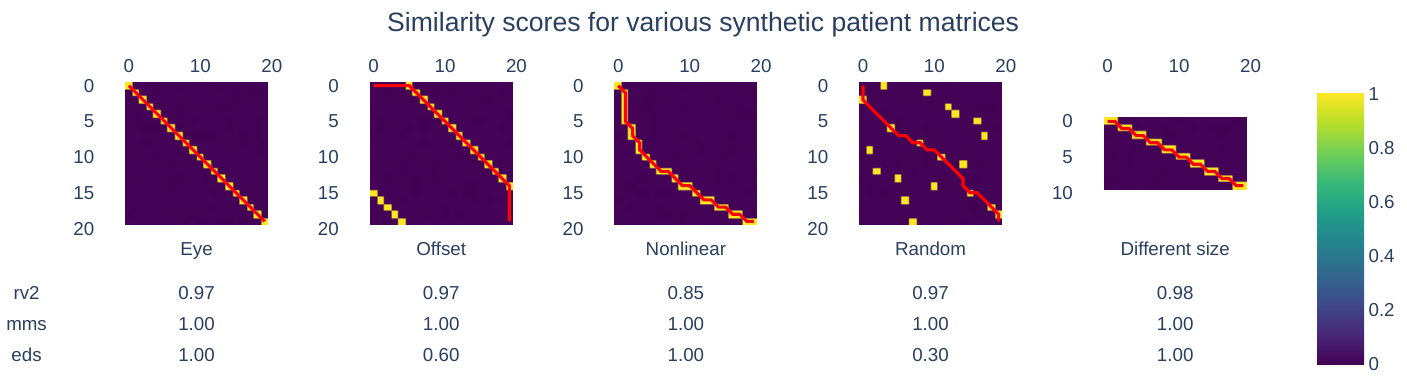}}
\caption{Comparison of matrix similarity methods of synthetic patient matrices. The underlying "patient" matrices were generated so that the similarity matrices look in a certain way. 
\textit{Matrices:} Similarity matrix of two synthetic patients.
\textit{Red lines:} the best path found by the \textit{eds} algorithm.
\textit{Table under matrices:} Similarity scores for given synthetic patient pair.~\cite{Zelina2023thesis}}\label{fig:matsim-synthetic}
\end{figure*}

Figure \ref{fig:matsim-synthetic} demonstrates the differences  of these methods on synthetic data.

\begin{itemize}
    \item \textbf{RV coefficient based similarity (\texttt{Rrv2})}\newline
    The first approach uses the RV coefficient~\cite{rv-coeff} -- an extension of the Pearson correlation coefficient for multivariate random variables. To be able to compare patients with a different number of notes, a modified version of the RV coefficient developed by Smilde et al.~\cite{rv2} is used. This ensures the similarity score is bounded between $-1$ and $1$.

    \textit{rv2} takes into account the temporal aspect of notes -- patients whose similarity matrix is mostly diagonal have higher rv2 similarity than those with the same values in their similarity matrix but distributed randomly.

    \item \textbf{MaxMax matrix similarity (\texttt{Rmms})}\newline
    This method finds the best possible note-to-note matching while discarding any temporal information. To calculate the similarity score for a pair of patients, \textit{mms} uses their pairwise cosine-similarity matrix. It first calculates the row-wise and column-wise maximum values ($M_r$ and $M_c$) and concatenates them into a single vector $M$. The resulting similarity score is the mean value of vector $M$. 

    In some categories, there are only a few relevant notes for each patient, and this method was designed to deal with such cases. Implementation details are available in~\cite{Zelina2023thesis}.

    \item \textbf{Edit-Distance matrix similarity (\texttt{Reds})}\newline
    This method finds the best path through the patient similarity matrix (the red paths in Figure \ref{fig:matsim-synthetic}). It is very similar to dynamic time warping~\cite{dtw}, which is often used, for example, when aligning speech data.

    The method uses a dynamic algorithm to find the path that has the greatest arithmetic mean of cosine similarities of the tiles (pairs of notes) it visits. The method finds the best time-consistent matching of notes from the first to last notes of both patients. Implementation details are available in~\cite{Zelina2023thesis}.
\end{itemize}

\section{Clinical notes dataset}

The clinical experts at Masaryk Memorial Cancer Institute assisted us with getting the electronic health records of some of their patients. The dataset contains clinical notes, records from medical examinations, and laboratory tests in the Czech language. The inclusion criteria were a breast cancer diagnosis (i.e., the C50* codes in ICD-10~\cite{icd10}) and consequent treatment recorded between January 2017 and December 2021. The exclusion criteria were the patients' opting out of the secondary research use of their medical data. The dataset was primarily acquired for a patient empowerment research project this work was a part of. The Ethics Committee of Masaryk Memorial Cancer Institute approved the use of the data in the project in May 2021 (the corresponding records can be provided on request).

In total, there are 152,552 records of 4,267 unique patients in the data. This yields around 36 records per patient on average. 
The dataset is in XML format, one file per patient, and includes useful metadata like datetime of record creation, patient ICD-10 diagnosis code~\cite{icd10}, or procedure identification in some laboratory tests. On the other hand, many fields, like patient name, are intentionally excluded or pseudonymized.

In terms of disk space, the XML files take up 363~MB, whereas the pure clinical notes text without metadata takes up only 300 MB. 

Due to privacy, ethical and legal concerns, we cannot share this dataset, but we can provide more details to interested parties on request.

\section{Experiments}
\subsubsection{Explored parameter space}

For the evaluation, we combine all previously described algorithms and their hyperparameters, which compound multiplicatively, leading to 42 experiments in total. The full list of tested combinations follows: 
\begin{itemize}
    \item 2 filtering options \newline
        \hspace*{10pt} {\small\texttt{filtered (\cmark), non-filtered (\xmark)}}
    \item 7 vectorization options \newline
        \hspace*{10pt} 3 algorithms $\times$ 2 dimensionality + 1 combination  \newline
        \hspace*{10pt} {\fontsize{8}{10}\texttt{Vlsa050, Vlsa200, Vd2v050, Vd2v200, Vrbc050, Vrbc200, combined}}
    \item 3 matrix similarity algorithms \newline
        \hspace*{10pt} {\small\texttt{Rrv2, Rmms, Reds}}
\end{itemize}

The \texttt{combined} vectorization option ensembles all three vectorization methods with dimensionality 50. First, it calculates similarity scores for each vectorization option separately. Afterwards, it takes the mean of the calculated scores and uses it as the output similarity score.

\subsubsection{Performance}
Table \ref{tab:matsim-speeds} compares how fast the matrix similarity methods are. The table shows how many minutes it takes to recalculate similarity scores for the whole dataset (4,267 patients, approx. 9 M pairwise scores). We can see that the \textit{rv2} method slows down considerably with increased dimension.

\begin{table}[h]
\caption[Time complexity of matrix similarity methods]{Time (in minutes) it takes to calculate the entire patient similarity matrix (4267 patients, approx. 9 M pairwise scores) with different methods. Measured on a machine with 12 CPU cores.}
\centering
\begin{tabular}{r|ccc}
dimension & rv2 & mms & eds \rule{0pt}{2.3ex}\\
\hline
50  & 3.5 & 15 & 140    \rule{0pt}{2.6ex}\\
200 & 280 & 50 & 400  \\    
\end{tabular}
\label{tab:matsim-speeds}
\end{table}

\section{Validation study}
To be able to compare different methods and parameters, we have designed a validation study. The ground-truth patient similarity annotations were provided by three clinicians from the Masaryk Memorial Cancer Institute (MMCI).\footnote{\weblink{https://www.mou.cz/en/}}

\subsubsection{Similarity categories}
\label{chap:val-categories}
Because the concept of patient similarity is not well defined,  we have developed a set of ten distinct similarity categories in collaboration with the clinicians from MMCI. The results are categories that are relevant to all the participating clinicians when making decisions about their patients.

The categories are as follows:

\begin{multicols}{2}
\begin{enumerate}
   \item Age
   \item Family history
   \item Medical history
   \item Social history
   \item Medication
   \item Allergies
   \item Type of tumor
   \item Treatment
   \item Treatment type
   \item Side effects
\end{enumerate}
\end{multicols}

The difference between \textit{Treatment} and \textit{Treatment type} is the following:

\textit{Treatment} is meant as the top-level treatment strategy, e.g., adjuvant therapy, neoadjuvant therapy or palliative care.

\textit{Treatment type} is meant as a concrete treatment approach, such as chemotherapy, radiation therapy, or surgery.

\subsection{Study design}
In cooperation with MMCI, we have chosen 10 \textit{pivot} patients that were deemed representative of the typical patients present in the entire dataset. For each \textit{pivot} patient, we have then selected 5 \textit{relevant} patients based on a baseline version of the similarity system (more details in the next two paragraphs)  -- 2 similar, 1 neutral, and 2 dissimilar.
For each \textit{pivot} patient, the annotators were tasked with evaluating how similar a given \textit{relevant} patient is to the \textit{pivot} patient in each of the 10 categories.

We selected the \textit{relevant} patients with a baseline system so that the relatively small number of annotations could cover at least some similar patients. Without it, we could have randomly selected 5 patients that are totally different from the \textit{pivot}, which would lead to degenerate annotations. The presentation order of the selected patients was randomized for each annotator.

The baseline system scored similarity based on \textit{lsa} embeddings  and cosine similarity of all concatenated patient records.


The evaluators used a 0~--~10 scale, where 0 is totally different, and 10 means totally similar. They could also assign a $-1$ when they deemed the two patients incomparable. In addition, we included space for a free-form comment for each pivot-relevant patient comparison pair, so that the annotators can indicate special circumstances or feedback.

At first, we wanted to enforce a strict ordering of the relevant patients (two different relevant patients cannot have the same score in the same category). This would simplify the evaluation, but in the end, we decided not to enforce it in the interest of simplifying the annotation process.

A completely filled-in annotation leads to 500 (10 pivots $\times$~5 relevant patients per pivot $\times$~10 categories) similarity scores for a single annotator.

Annotations were collected through a secure internal web tool. An in-depth description is available in~\cite{Zelina2023thesis}.



\section{Results}

\subsubsection{Evaluation methodology}
We use the Kendall rank correlation coefficient~\cite{kendall-book} to evaluate the patient similarity results. This measure allows us to compare two sequences of similarity scores as orderings -- looking only at their ranks and discarding the absolute differences. This way we do not have to worry about normalizing predicted scores or possible different scales individual annotators used.

The validation dataset is not very big and we always measure the correlation of only 5 samples (each pivot has 5 selected relevant patients). The annotations also often contain tied values. For this reason, we have chosen to use Kendall~$\tau$ as opposed to Spearman's~$\rho$. Kendall et al.~\cite{kendall-book} argue that Kendall~$\tau$ has more reliable confidence intervals than Spearman's~$\rho$. According to Bonett and Wright~\cite{kendall-samples} $\tau$ requires fewer samples than $\rho$ for the same statistical significance.
Kendall $\tau$ is also more robust than $\rho$~\cite{kendall-vs-spearman}.

\subsubsection{Result data}

\begin{table*}[tb]
\caption{Mean performance of each parameter combination. Numbers show the mean performance over all similarity categories. The performance in each category is measured as the mean of the Kendall rank correlation coefficients between mean annotations and model output scores. }
\centering
\setlength{\tabcolsep}{3pt}
\begin{tabular}{r|cc|cc|cc|cc|cc|cc|cc}
 & \multicolumn{2}{c|}{\texttt{combined}} & \multicolumn{2}{c|}{\texttt{Vd2v050}} & \multicolumn{2}{c|}{\texttt{Vd2v200}} & \multicolumn{2}{c|}{\texttt{Vlsa050}} & \multicolumn{2}{c|}{\texttt{Vlsa200}} & \multicolumn{2}{c|}{\texttt{Vrbc050}} & \multicolumn{2}{c}{\texttt{Vrbc200}} \\
filter & \xmark & \cmark & \xmark & \cmark & \xmark & \cmark & \xmark & \cmark & \xmark & \cmark & \xmark & \cmark & \xmark & \cmark \\
\hline
\texttt{Rrv2} & \colorize{7ac665}{000000}{0.24}  & \colorize{036e3a}{f1f1f1}{0.31}  & \colorize{73c264}{000000}{0.24}  & \colorize{91d068}{000000}{0.23}  & \colorize{0f8446}{f1f1f1}{0.30}  & \colorize{d5ed88}{000000}{0.19}  & \colorize{b1de71}{000000}{0.21}  & \colorize{006837}{f1f1f1}{0.31}  & \colorize{c9e881}{000000}{0.20}  & \colorize{06733d}{f1f1f1}{0.31}  & \colorize{b7e075}{000000}{0.21}  & \colorize{89cc67}{000000}{0.23}  & \colorize{bfe47a}{000000}{0.20}  & \colorize{89cc67}{000000}{0.23}  \\
\texttt{Rmms} & \colorize{0e8245}{f1f1f1}{0.30}  & \colorize{15904c}{f1f1f1}{0.29}  & \colorize{feeb9d}{000000}{0.14}  & \colorize{fffbb8}{000000}{0.15}  & \colorize{fa9857}{000000}{0.08}  & \colorize{fed683}{000000}{0.12}  & \colorize{87cb67}{000000}{0.24}  & \colorize{0c7f43}{f1f1f1}{0.30}  & \colorize{a0d669}{000000}{0.22}  & \colorize{16914d}{f1f1f1}{0.29}  & \colorize{73c264}{000000}{0.25}  & \colorize{ebf7a3}{000000}{0.17}  & \colorize{7ac665}{000000}{0.24}  & \colorize{ecf7a6}{000000}{0.17}  \\
\texttt{Reds} & \colorize{15904c}{f1f1f1}{0.29}  & \colorize{36a657}{f1f1f1}{0.27}  & \colorize{fdbb6c}{000000}{0.10}  & \colorize{f67c4a}{f1f1f1}{0.07}  & \colorize{b91326}{f1f1f1}{0.01}  & \colorize{ca2427}{f1f1f1}{0.02}  & \colorize{c7e77f}{000000}{0.20}  & \colorize{82c966}{000000}{0.24}  & \colorize{a5d86a}{000000}{0.22}  & \colorize{8ecf67}{000000}{0.23}  & \colorize{daf08d}{000000}{0.19}  & \colorize{ddf191}{000000}{0.18}  & \colorize{daf08d}{000000}{0.19}  & \colorize{d3ec87}{000000}{0.19}  \\
\end{tabular}

\label{tab:result-summary}
\end{table*}

\begin{table*}[t]
\caption{Detailed performance of the best 10 parameter combinations. Numbers show the mean of the Kendall rank correlation coefficients between mean annotations and model output scores in each similarity category. \textit{mmethod}: Patient matrix similarity method. \textit{vmethod}: Note vectorization method. \textit{filter}: category-specific similarity (\cmark, with segment filtering) or general similarity (\xmark, without segment filtering)}
\centering
\setlength{\tabcolsep}{2.5pt}
\begin{tabular}{ccc|cccccccccc|c}
\textit{mmethod} & \textit{vmethod} & \textit{filter} & 01 & 02 & 03 & 04 & 05 & 06 & 07 & 08 & 09 & 10 & mean \\
\hline
\texttt{Rrv2} & \texttt{Vlsa050} & \cmark & \colorize{f57748}{f1f1f1}{0.15}  & \colorize{279f53}{f1f1f1}{0.62}  & \colorize{f67c4a}{f1f1f1}{0.16}  & \colorize{c62027}{f1f1f1}{0.05}  & \colorize{f57245}{f1f1f1}{0.15}  & \colorize{fffab6}{000000}{0.34}  & \colorize{f2faae}{000000}{0.37}  & \colorize{b5df74}{000000}{0.47}  & \colorize{249d53}{f1f1f1}{0.62}  & \colorize{fdbb6c}{000000}{0.23}  & \colorize{fff0a6}{000000}{\textbf{0.31}}  \\
\texttt{Rrv2} & \texttt{combined} & \cmark & \colorize{fdbf6f}{000000}{0.23}  & \colorize{a5d86a}{000000}{0.49}  & \colorize{de402e}{f1f1f1}{0.09}  & \colorize{a50026}{f1f1f1}{0.00}  & \colorize{db382b}{f1f1f1}{0.08}  & \colorize{fed884}{000000}{0.27}  & \colorize{fee593}{000000}{0.29}  & \colorize{93d168}{000000}{0.51}  & \colorize{07753e}{f1f1f1}{0.68}  & \colorize{b5df74}{000000}{0.47}  & \colorize{feeda1}{000000}{\textbf{0.31}}  \\
\texttt{Rrv2} & \texttt{Vlsa200} & \cmark & \colorize{fb9d59}{000000}{0.19}  & \colorize{78c565}{000000}{0.54}  & \colorize{f67c4a}{f1f1f1}{0.16}  & \colorize{b71126}{f1f1f1}{0.03}  & \colorize{ec5c3b}{f1f1f1}{0.12}  & \colorize{fdc372}{000000}{0.24}  & \colorize{e8f59f}{000000}{0.39}  & \colorize{82c966}{000000}{0.53}  & \colorize{249d53}{f1f1f1}{0.62}  & \colorize{feca79}{000000}{0.25}  & \colorize{feec9f}{000000}{\textbf{0.31}}  \\
\texttt{Rmms} & \texttt{Vlsa050} & \cmark & \colorize{f57748}{f1f1f1}{0.15}  & \colorize{b5df74}{000000}{0.47}  & \colorize{ea5739}{f1f1f1}{0.12}  & \colorize{e34933}{f1f1f1}{0.10}  & \colorize{db382b}{f1f1f1}{0.08}  & \colorize{feda86}{000000}{0.27}  & \colorize{dcf08f}{000000}{0.41}  & \colorize{d5ed88}{000000}{0.42}  & \colorize{3ca959}{f1f1f1}{0.60}  & \colorize{f5fbb2}{000000}{0.37}  & \colorize{fee999}{000000}{0.30}  \\
\texttt{Rmms} & \texttt{combined} & \xmark & \colorize{fdaf62}{000000}{0.21}  & \colorize{fdb163}{000000}{0.21}  & \colorize{fed27f}{000000}{0.26}  & \colorize{f47044}{f1f1f1}{0.14}  & \colorize{f36b42}{f1f1f1}{0.14}  & \colorize{a90426}{f1f1f1}{0.01}  & \colorize{7dc765}{000000}{0.53}  & \colorize{93d168}{000000}{0.51}  & \colorize{249d53}{f1f1f1}{0.62}  & \colorize{fff5ae}{000000}{0.33}  & \colorize{fee797}{000000}{0.30}  \\
\texttt{Rrv2} & \texttt{Vd2v200} & \xmark & \colorize{fecc7b}{000000}{0.25}  & \colorize{fa9857}{000000}{0.19}  & \colorize{f99355}{000000}{0.18}  & \colorize{f7814c}{f1f1f1}{0.16}  & \colorize{e95538}{f1f1f1}{0.11}  & \colorize{f57748}{f1f1f1}{0.15}  & \colorize{91d068}{000000}{0.51}  & \colorize{7fc866}{000000}{0.53}  & \colorize{51b35e}{f1f1f1}{0.58}  & \colorize{fee491}{000000}{0.29}  & \colorize{fee797}{000000}{0.30}  \\
\texttt{Rmms} & \texttt{combined} & \cmark & \colorize{fdbd6d}{000000}{0.23}  & \colorize{e9f6a1}{000000}{0.39}  & \colorize{f67c4a}{f1f1f1}{0.16}  & \colorize{a50026}{f1f1f1}{0.00}  & \colorize{e44c34}{f1f1f1}{0.10}  & \colorize{f98e52}{f1f1f1}{0.18}  & \colorize{f2faae}{000000}{0.37}  & \colorize{b5df74}{000000}{0.47}  & \colorize{3ca959}{f1f1f1}{0.60}  & \colorize{ebf7a3}{000000}{0.39}  & \colorize{fee491}{000000}{0.29}  \\
\texttt{Reds} & \texttt{combined} & \xmark & \colorize{fdaf62}{000000}{0.21}  & \colorize{fa9857}{000000}{0.19}  & \colorize{fdb567}{000000}{0.22}  & \colorize{fa9656}{000000}{0.18}  & \colorize{d93429}{f1f1f1}{0.08}  & \colorize{e24731}{f1f1f1}{0.10}  & \colorize{f2faae}{000000}{0.37}  & \colorize{6ec064}{000000}{0.55}  & \colorize{249d53}{f1f1f1}{0.62}  & \colorize{f5fbb2}{000000}{0.37}  & \colorize{fee491}{000000}{0.29}  \\
\texttt{Rmms} & \texttt{Vlsa200} & \cmark & \colorize{ef633f}{f1f1f1}{0.13}  & \colorize{f5fbb2}{000000}{0.37}  & \colorize{f7814c}{f1f1f1}{0.16}  & \colorize{c82227}{f1f1f1}{0.05}  & \colorize{f57245}{f1f1f1}{0.15}  & \colorize{fdb96a}{000000}{0.23}  & \colorize{cfeb85}{000000}{0.43}  & \colorize{b5df74}{000000}{0.47}  & \colorize{51b35e}{f1f1f1}{0.58}  & \colorize{feeda1}{000000}{0.31}  & \colorize{fee491}{000000}{0.29}  \\
\texttt{Reds} & \texttt{combined} & \cmark & \colorize{f16640}{f1f1f1}{0.13}  & \colorize{fee491}{000000}{0.29}  & \colorize{fca55d}{000000}{0.20}  & \colorize{a50026}{f1f1f1}{-0.08}  & \colorize{db382b}{f1f1f1}{0.08}  & \colorize{fdb768}{000000}{0.22}  & \colorize{e8f59f}{000000}{0.39}  & \colorize{c5e67e}{000000}{0.45}  & \colorize{69be63}{f1f1f1}{0.56}  & \colorize{b5df74}{000000}{0.47}  & \colorize{feda86}{000000}{0.27}  \\
\end{tabular}
\label{tab:results-top10}
\end{table*}

\begin{figure*}[t]%
\centering
{\includegraphics[width=\textwidth]{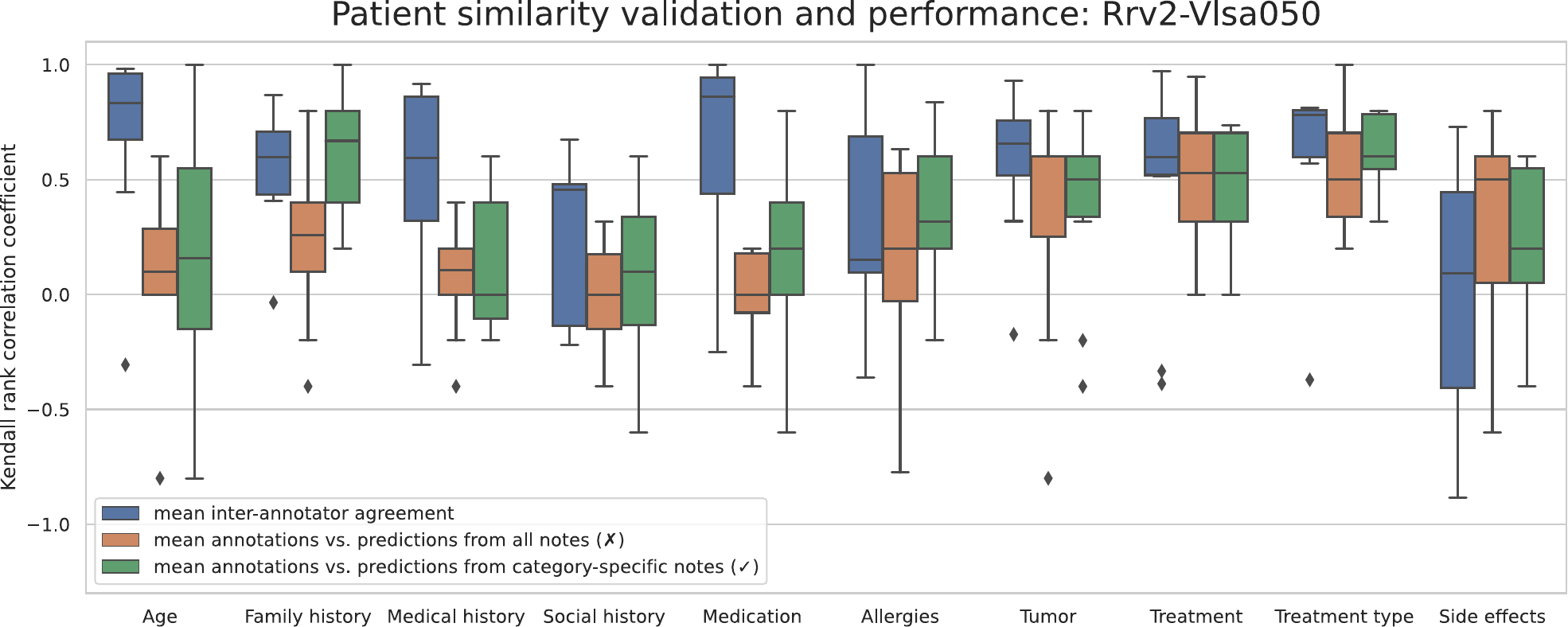}}
\caption{Detailed performance analysis of the \texttt{Rrv2-Vlsa050} combination (best performing). The boxplot shows a comparison of the inter-annotator agreement (\textit{blue}), the performance of the model without filtering (\textit{orange}, \xmark), and the performance of the model with filtering (\textit{green}, \cmark) for different categories.}\label{fig:results-bars}
\end{figure*}

Table \ref{tab:result-summary} summarises the performance of all 42 parameter combinations. Green cell background means a higher correlation, red background means a lower correlation. We have observed the following:
\begin{itemize}
    \item Filtering improves the performance, mainly for the \texttt{Vlsa} and \texttt{combined} vectorization techniques.
    \item \texttt{Rrv2} seems to work well with all vectorization techniques.
    \item The \texttt{combined} vectorization ensemble has a stabilizing effect and in the case of \texttt{Reds} improves the performance significantly.
    \item The best option is the \texttt{Rrv2} matrix similarity method with either the \texttt{Vlsa} or \texttt{combined} vectorization.
    \item \texttt{Vd2v} performs poorly, especially in combination with \texttt{Rmms} or \texttt{Reds}.
\end{itemize}

The low performance of \texttt{Vd2v200} with filtering is most likely caused by insufficient training data for the Doc2Vec algorithm with dimension 200. In some similarity categories, most note segments are irrelevant and discarded, and thus, the amount of text is low.  We can see that lowering the dimensionality helps (\texttt{Vd2v050}), but the performance does not surpass the non-filtered version. 

Table \ref{tab:results-top10} shows the detailed performance of the 10 best parameter combinations, including the per-category correlations. We can see that the performance in different categories varies greatly. We explore the possible reasons in the discussion section.

Figure \ref{fig:results-bars} shows a boxplot of inter-annotator agreement as well as the results of the best-performing model (both filtered and unfiltered). We can see that in some categories, namely \textit{social history}, \textit{allergies}, and \textit{side effects}, the agreement is poor and thus the results for these categories are unreliable. We elaborate on this important observation in the discussion section.

\subsubsection{Effects of vectorization hyperparameters}
In order to analyze the effects of other hyperparameters that were not part of the main grid search, we ran several experiments for each vectorization technique. We concluded that most vectorization hyperparameters have a minimal effect on the final performance. The only exception is finetunig of the RobeCzech model, which greatly improves the performance.

Detailed results of these experiments are available in \cite{Zelina2023thesis}.

\section{Discussion}

\subsubsection{Performance of different similarity categories.}

From Table \ref{tab:results-top10} and Figure \ref{fig:results-bars}, it is evident that predictions for 6 out of the 10 similarity categories are mostly random (categories 01,  03 to 06, and 10).  We have inspected various intermediate outputs of the systems and cross-referenced them with the clinical notes and the validation annotations, and we offer the following commentary:
\begin{itemize}
    \item Category \textit{01:~Age}: age information is not present in the clinical note texts.

    \item Categories \textit{04:~Social history}, \textit{06:~Allergies}, and \textit{10:~Side effects} have a quite low annotator agreement, which may be a sign that they are not well defined or that the annotators approached them differently. We also noticed that there is either little information about these categories present in the clinical notes, or it is mentioned only once. This might also be a part of the problem, which we plan to revisit in the near future by preparing a more focused benchmark dataset with the clinicians.

    \item Some categories have a low segment count, as the notes contain only a few relevant pieces of information. This is the case for category-specific predictions of categories 01, 02, 04, 06, and 10.
    \begin{itemize}
        \item Unlike more represented categories, they are very sensitive to individual segments. 
        \item Similarity matrices are often dominated by short notes with only a few segments and very similar text. These texts are usually not as relevant for the category as the system believes. (e.g. \textit{patient lives alone}, or \textit{Started taking medicine on 26. January 2020})
        \item This could be remedied by being more strict when selecting relevant titles for categories (in the filtering step), but this means that more patients would have fewer segments or be completely filtered out.
    \end{itemize}

    \item Categories \textit{03:~Medical history} and \textit{05:~Medication} have a reasonable number of notes, but these notes have fewer and shorter segments than notes for \textit{07:~Type of tumor} or \textit{08:~Treatment}. Thus, they suffer from similar deficiencies as the low-note-count categories.

\end{itemize}

On the other hand, considering the system is not trained on the validation similarities in any way, the scores for the categories \textit{07:~Type of tumor}, \textit{08:~Treatment}, and \textit{09:~Treatment type} perform reasonably well and the corresponding patient similarity rankings are comparable to the gold-standard annotations.

This makes sense, as the main focus of clinical notes is to document the treatment process. The whole experimental framework has also been built to analyze data changing over time. The treatment notes are more representative of this as opposed to family, medical or social history.

\subsubsection{Validation annotations}
In order to be able to evaluate the system performance, we designed a validation study in cooperation with MMCI. We have been able to gather three sets of similarity annotations for 50 pairs of patients in 10 similarity categories. The annotations were provided by three clinicians from MMCI.

We are thankful to all parties that made this study possible; however, during the evaluation, we have since discovered some limitations of the current setup. We validate the ordering of 5 patients based on their similarity to a pivot patient. This is rather different from the simple binary classification (similar/dissimilar) used in related studies (e.g.~\cite{Sun2012-pacsim}). Therefore, our results (including the annotator agreement values) lack robust statistical significance.

In the future, we would like to include more pivot patients as well as more than 5 comparison patients, which would make the study more statistically robust. Also, it would enable us to split the resulting dataset into train and test parts, thus allowing us to use supervised ML methods for patient similarity calculations. We would also like to refine the specification of some categories based on their broader clinical context to improve the inter-annotator agreement.

\section{Conclusion}
We have introduced several methods for calculating patient similarity from unstructured clinical notes and compared their performance on a validation dataset. 
The main evaluation consisted of a grid search of 42 combinations of filtering (2), vectorization methods (7), and patient matrix similarity methods (3).

We have shown that certain combinations of the investigated techniques perform well when predicting patient similarities based on the \textit{07:~Type of tumor}, \textit{08:~Treatment}, and \textit{09:~Treatment type} (and to some degree \textit{02: Family history}) aspects. Our system is not able to capture the other similarity aspects.

The best performing and stable option is the \textit{combined} embedding and the \textit{Rv2} similarity method. This combination can thus be recommended for computing patient similarities in relevant downstream tasks.

\begin{credits}

\subsubsection{\ackname} Supported by the Grant Agency of Masaryk University under grant no. MUNI/G/1763/2020 (AIcope). Furthermore, this project has received funding from the European Union’s Horizon research and innovation programme under grant agreement no. 101057048, the project SALVAGE (OP JAK; reg. no. CZ.02.01.01/00/22\_008/0004644, co-funded by the European Union and by the State Budget of the Czech Republic) and the project DecideHeatlh, funded by the Technology Agency of the Czech Republic under the grant number TQ12000018. Computational resources were provided by the e-INFRA CZ project (ID:90254), supported by the Ministry of Education, Youth and Sports of the Czech Republic.

\end{credits}
%
%
%
\bibliographystyle{splncs04}
\bibliography{cas-refs}
%

\end{document}